\documentclass[10pt,twocolumn,letterpaper]{article}

\newcommand{\etal}{\emph{et al. }}

\newcommand{\bx}{\mathbf{x}}

\newcommand{\by}{\mathbf{y}}

\newcommand{\bM}{\mathbf{M}}

\newcommand{\bS}{\mathbf{S}}

\newcommand{\bX}{\mathbf{X}}
\newcommand{\bY}{\mathbf{Y}}

\newcommand{\hx}{\hat{x}}

\newcommand{\bhy}{\hat{\mathbf{y}}}

\newcommand{\bhY}{\hat{\mathbf{Y}}}

\newcommand{\wx}{\widetilde{\mathbf{x}}}

\newcommand{\mF}{\mathcal{F}}

\newcommand{\mJ}{\mathcal{J}}

\newcommand{\mM}{\mathcal{M}}

\newcommand{\mL}{\mathcal{L}}

\usepackage[table]{xcolor}
\usepackage{tikz}  %
\usepackage{iccv}
\usepackage{times}
\usepackage{epsfig}
\usepackage{graphicx}
\usepackage{amsmath}
\usepackage{amssymb}
\usepackage{subcaption}

\usepackage{subfiles}
\usepackage{pgfplots} %
\usepackage{pgfplotstable} %
\pgfplotsset{compat=newest} %
\usetikzlibrary{plotmarks} %
\usetikzlibrary{colorbrewer} %
\usepackage{booktabs} %
\usepackage{etoolbox,siunitx} %
\usepackage{caption}
\robustify\bfseries
\robustify\itshape
\usepgfplotslibrary{statistics}

\definecolor{orange}{cmyk}{0,0.6,1,0}
\definecolor{ForestGreen}{rgb}{0.0, 0.5, 0.0}

\newcommand{\ours}{FaSTGAN}
\newcommand{\nours}{FaSTGAN}
\newcommand{\nnours}{Ours}

\newcommand{\npremvos}{PReMVOS}

\newcommand{\nosvoss}{OSVOS$^\textrm{S}$}
\newcommand{\npml}{PML}

\newcommand{\nrgmp}{RGMP}

\newcommand{\nmonet}{MoNet}
\newcommand{\nvideomatch}{VideoMatch}
\newcommand{\nmaskrnn}{MaskRNN}
\newcommand{\ncrn}{CRN}
\newcommand{\ndyenet}{DyeNet}
\newcommand{\nosmn}{OSMN}
\newcommand{\nytvos}{YTVOS}

\newcommand{\nbvs}{BVS}

\definecolor{rowblue}{RGB}{220,230,240}

\usepackage[breaklinks=true,bookmarks=false]{hyperref}

\iccvfinalcopy %

\def\iccvPaperID{****} %
\def\httilde{\mbox{\tt\raisebox{-.5ex}{\symbol{126}}}}

\begin{document}

\title{Fast video object segmentation with Spatio-Temporal GANs}

\author{S. Caelles$^{1,}$\thanks{First two authors contributed equally} \quad
        A. Pumarola$^{2,*}$ \quad
        F. Moreno-Noguer$^{2}$ \quad
        A. Sanfeliu$^{2}$ \quad
        L. Van Gool$^{1}$ \\
$^1$Computer Vision Lab, ETH Z\"urich \quad
$^2$Institut de Robotica i Inform\`atica Industrial, CSIC-UPC \\
}

\maketitle

\iccvfinalcopy %

\def\iccvPaperID{****} %
\def\httilde{\mbox{\tt\raisebox{-.5ex}{\symbol{126}}}}

\ificcvfinal\pagestyle{empty}\fi
\setcounter{page}{1}
\begin{abstract}
\label{sec:abstract}
Learning descriptive spatio-temporal object models from data is paramount for the task of semi-supervised video object segmentation. 
Most existing approaches mainly rely on models that estimate the segmentation mask based on a reference mask at the first frame (aided sometimes by optical flow or the previous mask). 
These models, however, are prone to fail under rapid appearance changes or occlusions due to their limitations in modelling the temporal component.
On the other hand, very recently, other approaches learned long-term features using a  convolutional LSTM to leverage the information from all previous video frames. 
Even though these models achieve better temporal representations, they still have to be fine-tuned for every new video sequence. 
In this paper, we present an intermediate solution and devise a novel GAN architecture, \nours{}, to learn spatio-temporal object models over finite temporal windows. 
To achieve this, we concentrate all the heavy computational load to the training phase with two critics that enforce spatial and temporal mask consistency over the last $K$ frames. Then at test time, we only use a relatively light regressor, which reduces the inference time considerably. 
As a result, our approach combines a high resiliency to sudden geometric and photometric object changes with efficiency at test time (no need for fine-tuning nor post-processing).
We demonstrate that the accuracy of our method is on par with state-of-the-art techniques on the challenging YouTube-VOS and DAVIS datasets, while running at 32 fps, about  $4\times$ faster than the closest competitor.
\end{abstract}

\section{Introduction}
\label{sec:intro}

The problem of semi-supervised video object segmentation consists in segmenting an object from the background throughout a video sequence given its ground truth mask in the initial frame. Large video datasets like DAVIS~\cite{Perazzi2016,Pont-Tuset_arXiv_2017} and the recently released YouTube-VOS~\cite{xu2018youtube} have spurred a number of deep networks methods~\cite{caelles2017one,Perazzi2017,voigtlaender2017online,Man+18b,Cheng_ICCV_2017,fusionseg,Hu_2018_CVPR,Xiao_2018_CVPR,Yoon_2017_ICCV,Yang_2018_CVPR,xu2018youtube,wug2018fast,Chen_2018_CVPR,Ci_2018_ECCV,Hu_2018_ECCV,Bao_2018_CVPR,Chandra_2018_CVPR,Li_2018_ECCV,luiten2018premvos} that improve the performance of approaches from the pre-deep learning era~\cite{Chang2013,Grundmann2010,Maerki_CVPR_2016,Perazzi_2015_ICCV,Tsai2012motion} by a large margin. The problem, however, is still far from being solved. Occlusions, rapid object movements, appearance changes and similarity among different instances are still a major obstacle that often require heavy post-processing operations, human intervention and expensive model fine-tuning.

In order to achieve robustness to these challenges, descriptive spatio-temporal object models encoding appearance and geometric changes need to be learned.  
Most existing state-of-the-art approaches~\cite{caelles2017one,Perazzi2017,Yang_2018_CVPR} heavily rely on a reference mask to fine-tune the model and in some cases, use the previous mask as a guidance.
Formally, if we denote this reference mask by $\bY_0$ and the RGB frame at time $t$ by $\bX_t$, these approaches model the mask $\bhY_t$ as $p(\bhY_t|\bY_0,\bX_0,\bX_{t-1},\bX_t)$ or $p(\bhY_t|\bY_0,\bX_0,\bX_{t-1},\bX_t, \bhY_{t-1})$. 
Since no temporal consistency is enforced, these methods tend to be robust to drifting, but they underperform when the object drastically changes its appearance. 

This can be remedied by leveraging the  temporal consistency of the segmented mask. However, while this  was a common practice in the past~\cite{Grundmann2010,Maerki_CVPR_2016,Perazzi_2015_ICCV}, it is not usual among deep learning methods, in part due to the absence of large scale video object segmentation datasets. Very recently, Xu \etal~\cite{xu2018youtube} used a convolutional LSTM trained with the YouTube-VOS dataset to learn long-term temporal dependencies from the entire history of the object in the video. That is, the mask $\bhY_t$ is modeled as $p(\bhY_t|\bY_0,\bX_0,\bX_1,\ldots,\bX_t)$. While this approach demonstrates improved performance compared to previous baselines which did not enforce temporal consistency, it seems to be too generic, as it still needs a computationally demanding  fine-tuning step when applied to a sequence with unseen objects. 

\begin{figure*}[t!]
	\centering
	\includegraphics[width=\textwidth]{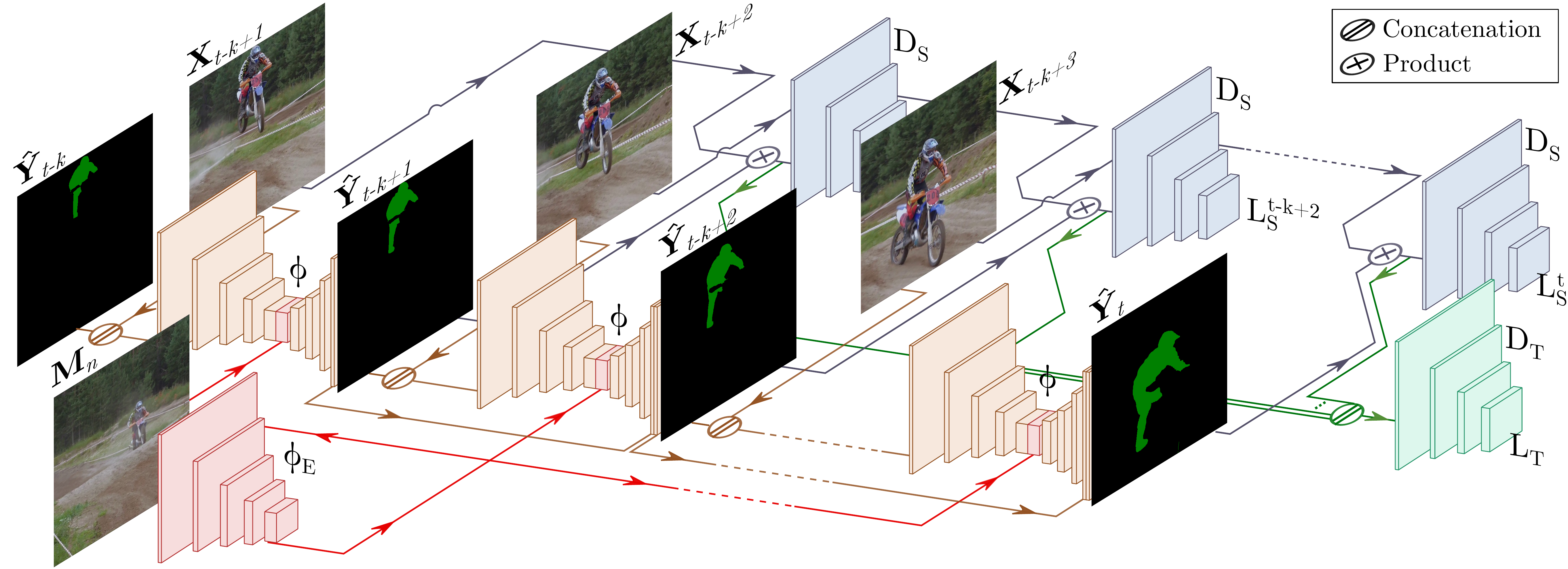}
	\caption{{\bf Overview of our method, \nours{}, for video object segmentation.} The diagram displays the model at training time for frames ${t-K+1, t-K+2, \ldots, t}$. The proposed architecture consists of three main components: a segmentation regressor $\phi$, a spatial critic $D_\text{S}$ and a temporal critic $D_\text{T}$. Weights are shared across each network of the same instance. $\phi_\text{E}$ is the encoder part of $\phi$ and it encodes the image-mask pair $\langle \bX_t, \bhY_{t-1} \rangle$ and $\bM_n$ at every time step, see Section~\ref{sec:model_architecture}.}
	\label{fig:overal_model}
	\vspace{-4mm}
\end{figure*}

In this paper, we propose \nours{}, an intermediate solution that learns spatio-temporal object appearance models over finite time horizons that does not require fine-tune nor post-processing. Essentially during training, we model the segmentation masks as $p(\bhY_t|\bY_0,\bX_0,\bX_{t-K},\ldots,\bX_t)$, where $K$ is the size of the temporal window. In order to implement this model, we design a regressor network architecture inspired by the agile Siamese encoder-decoder structure proposed by Wug \etal~\cite{wug2018fast}. In its original form, this regressor  is only fed by the reference mask and the masked image at the previous time step. To exploit all information within a temporal window of size $K$, we could naively make the regressor have access to more information by concatenating features from the $K$ previous frames. This, however, would heavily penalize the efficiency and adaptability of the model. We have therefore devised a novel GAN architecture (Figure~\ref{fig:overal_model}) in which, during training, this regressor is combined with $K+1$ discriminators that enforce the temporal and spatial coherence of the generated masks over the temporal window. At test, these discriminators are removed, keeping the original efficiency of the Siamese regressor, while allowing it to model the object across longer time horizons.

As a result, our  architecture only uses video data to train and does not require any kind of fine-tuning nor post-processing operations at test time. This makes our approach very efficient, running at 32 fps on $512\times 512$ video frames which is about  $4\times$ faster than~\cite{wug2018fast}, which was the fastest video segmentation method so far with $7.7$ fps reported in their original work. Furthermore, the accuracy of the segmentation masks we obtain on both DAVIS and YouTube-VOS datasets is on a par with state-of-the-art methods that focus on speed. All code, pre-trained models and pre-computed results used in this paper will be released.

\section{Related Work}
\label{sec:related}

\vspace{1mm}
\noindent\textbf{Video object segmentation:} In recent years, video object segmentation has experienced a tremendous increase in popularity due to the publication of large datasets (DAVIS~\cite{Perazzi2016, Pont-Tuset_arXiv_2017} and YouTube-VOS~\cite{xu2018youtube}) that have enabled the training of deep learning techniques. 
The two main settings to tackle this problem are semi-supervised and unsupervised. 
In the former, the ground truth mask for the object of interest in the first frame of the video sequence is given to the method whereas in the latter no information is given to the algorithm and usually the object with predominant motion is segmented. In this work, we tackle the semi-supervised setting and therefore we focus on previous work in that field.

Traditional approaches used temporal super-pixel~\cite{Chang2013, Grundmann2010}, optimization in the bilateral space~\cite{Maerki_CVPR_2016}, or optimal selection of object proposals~\cite{Perazzi_2015_ICCV} to obtain the object segmentation mask for each frame in the video sequence. \cite{caelles2017one}~and~\cite{Perazzi2017} were the first two approaches to apply deep learning to the problem. Specifically, 
~\cite{caelles2017one}~fine-tuned the network using the first frame of the video sequence whereas~\cite{Perazzi2017} used the mask from the previous frame as an input to the network. 
\cite{voigtlaender2017online}~extended~\cite{caelles2017one} with an online learning strategy, while~\cite{Man+18b} also extended~\cite{caelles2017one} by combining its result with Mask-RCNN~\cite{He_2017_ICCV}.

Another set of techniques have tried to incorporate the information of the first frame in different ways. 
\cite{Yoon_2017_ICCV}~formulated the problem as a pattern matching with the initial mask,~\cite{Yang_2018_CVPR} introduced the initial mask in the network in a batch-norm like layer. \cite{wug2018fast}~used a Siamese network to combine the low level features of the initial frame with the current one together with the back propagation through time training strategy introduced in~\cite{hu2017maskrnn}.

Moreover,  some methods have tried to leverage metric learning to solve the problem~\cite{Chen_2018_CVPR, Ci_2018_ECCV, Hu_2018_ECCV}, divide the object in multiple parts and track each of them~\cite{Cheng_2018_CVPR}, integrate CNN features with traditional energy minimization techniques~\cite{Bao_2018_CVPR, Chandra_2018_CVPR} or design a complex architecture with re-identification and bidirectional propagation modules~\cite{Li_2018_ECCV}.

Previous approaches mostly rely on single image segmentation, using at most the previous mask as a temporal consistency constraint. There are, however, a few attempts to exploit the temporal dimension better. For instance, some techniques have leveraged optical flow as an additional input to their network. \cite{Cheng_ICCV_2017}~and~\cite{fusionseg} built a second CNN branch to process optical flow, ~\cite{Hu_2018_CVPR} used it as a prior in the decoder of the network and, \cite{Xiao_2018_CVPR} used it to align the features from previous frames. While these methods use optical flow priors trained in a separate context, ~\cite{Xiao_2018_CVPR} used the large scale YouTube-VOS dataset (released by them) to train a convolutional LSTM in an end-to-end manner. 

Our approach lies in between methods that do not use temporal consistency, and~\cite{Xiao_2018_CVPR}, which learns long-term dependencies with an RNN.

\vspace{1mm}
\noindent\textbf{Generative Adversarial Networks:} Since the GAN framework was introduced in~\cite{goodfellow2014generative} to generate synthetic images from noise, its performance has drastically improved in subsequent works~\cite{radford2015unsupervised, karras2018progressive, brock2018large} achieving almost indistinguishable results from real images.
Moreover, it has been successfully applied in a wide range of different task such as conditional image synthesis~\cite{pix2pix2017, pumarola2018unsupervised}, video generation~\cite{pumarola2018ganimation, wang2018vid2vid}, domain adaptation~\cite{tsai2018adapt}, super-resolution~\cite{tensorlayer2017, LapSRN} or object detection~\cite{Wang_2017_CVPR}.

Recently, several works have tried using GANs for semantic segmentation~\cite{luc2016semantic, souly2017semi, luo2018macro, hung2018semiseg}.
\cite{luc2016semantic} trained the discriminator to differentiate between real and predicted probability maps.
\cite{luo2018macro} used two different discriminators to obtain local and global semantic consistency for the human part segmentation problem. 
~\cite{hung2018semiseg} tackled the semi-supervised semantic segmentation task using the discriminator to obtain a confidence label map for unlabeled data.

Finally, there have been several applications of GANs in video mainly for video synthesis~\cite{vondrick2016, tulyakov2017mocogan, saito2017tgan} and conditional video synthesis~\cite{wang2018vid2vid}. 
In video synthesis, \cite{saito2017tgan} splits the synthesis between two generators, the temporal generator that outputs a latent variable for each frame and the spatial generator that generates the frame from the latent variables.
\cite{tulyakov2017mocogan} splits the latent space in a motion subspace and a content subspace to generate videos with the same object but performing different motions.
Recently, in the conditional setting, \cite{wang2018vid2vid} synthesizes videos given the dense pose estimation, the semantic maps or the boundaries for each frame in the video sequence.

To the best of our knowledge, we are the first ones to successfully apply GANs to perform \textit{segmentation in videos}. 
Furthermore, we use a considerably higher image resolution than previous segmentation works (512 vs.\ 256 in~\cite{luo2018macro} or 321 in~\cite{hung2018semiseg}) and we leverage recent advances introduced in the images synthesis task, i.e.,\ WGAN with gradient penalty~\cite{gulrajani2017improved}, in order to improve stability during training.

\section{Problem Formulation}
\vspace{-1mm}
We next formally describe our problem, and generalize the formulation introduced in Section~\ref{sec:intro} to an arbitrary number of objects, i.e.,\ we aim  to design a deep learning  model able to segment and track  objects along a video sequence given only  one single segmentation mask per object. 

Let $\bx=(\bX_1,\ldots,\bX_{T})$ be an input RGB video with $T$ frames, where $\bX_t \in \mathbb{R}^{H \times W \times 3}$ denotes the \textit{t}\textsuperscript{th} frame. Let us also define $\mathbf{m}=(\bM_1,\ldots,\bM_{N})$ as a set of reference segmentations of $N$ objects. The reference segmentation $\bM_n \in \mathbb{R}^{H \times W \times (3+1)}$ for the $n$-th object is the concatenation of the first RGB frame in which the object appears with its annotated binary mask. Our goal is to estimate the masks $\bhy$ of all $N$ objects along the entire sequence $\bx$, i.e.,\  we want to learn the mapping $\mM: (\bx,\mathbf{m}) \rightarrow \bhy$, where $\bhy=(\bhY_1,\ldots,\bhY_{T})$, and  $\bhY_t \in \mathbb{R}^{H \times W \times N}$ contains the $N$ tracked objects masks in the \textit{t}\textsuperscript{th} video frame. We define the ground truth masks  $\by$ for a certain sequence $\bx$ as $\by=(\bY_1,\ldots,\bY_{T})$ .

\vspace{-1mm}
\section{Our Approach}
Figure~\ref{fig:overal_model} shows an overview of \nours{}, our proposed approach for video object segmentation. A regressor $\phi$ is trained on the binary segmentation task of separating the desired object from the background and two WGAN-GP~\cite{gulrajani2017improved} based critics, $D_\text{S}$ and $D_\text{T}$, enforce the model to produce semantically and temporally consistent estimates. To simplify the model, we introduce a Markov assumption defining the conditional distribution $p(\bhy| \bx,\mathbf{m})$ to be factorized as:
\begin{align}
p(\bhy| \bx,\mathbf{m}) = \prod_{t=1}^T p(\bhY_t| \bx_{t-K}^{t}, \mathbf{m}, \bhy_{t-K}^{t-1}),
\end{align}
meaning that we assume objects in a certain frame to be trackable given the reference segmentations $\mathbf{m}$, the current and $K$-$1$ previous frames $\bx_{t-K}^{t}=(\bX_{t-K},\ldots,\bX_{t})$, and the $K$-$1$ previous estimated segmentation masks $\bhy_{t-K}^{t-1}=(\bhY_{t-K},\ldots,\bhY_{t-1})$ . 

During training, the regressor $\phi$ is required to learn the mapping $\mM$ by modeling the distribution $p(\bhY_t| \bx_{t-K}^t, \mathbf{m}, \bhy_{t-K}^{t-1})$, as $\bhY_t = \phi(\bX_t,\mathbf{m}, \bhY_{t-1})$. A key property of our design is that the regressor does not directly receive the full information of the $K$-temporal window. Instead, it is trained to binary segment the $K$ frames, one at a time, given a single foreground/background segmentation $\bM_n$ of the desired object $(\bx_{t-K}^{t}, \bM_n) \rightarrow \bhy_{t-K}^{t}$. Each predicted mask  is then independently evaluated by a spatial critic $D_\text{S}(\bX_i, \bhY_i)$ $\forall i \in [t-K, t]$ that aims to penalize non-consistent semantic masks. The  temporal consistency is assessed by a temporal critic $D_\text{T}(\bx_{t-K}^{t}, \bhy_{t-K}^{t})$ that jointly evaluates the $K$ segmentation masks. Additionally, to feed the regressor with information from previous estimations we introduce a ``temporal skip connection" (see details in the following subsection). Note that with this strategy, the regressor is adapted to produce temporally coherent masks within a horizon of size $K$, without having to  simultaneously process the $K$ frames. This will be crucial to deliver a very fast regressor at test time, when the critics will be discarded. 

In the following subsections we describe in detail each of these components as well as the proposed training loss.

\subsection{Model Architecture}
\label{sec:model_architecture}
\paragraph{Segmentation Regressor.} Given the current frame $\bX_t$, the single-view reference segmentation $\bM_n$ of the desired object, and the previous estimate $\bhY_{t-1}$, the segmentation regressor $\phi$ aims to separate the desired object from the background producing the current estimate mask $\bhY_{t}=\phi(\bX_t, \bM_n, \bhY_{t-1})$. We denote the encoder part of $\phi$ as $\phi_\text{E}$. Similar to~\cite{wug2018fast}, $\phi_\text{E}$ maps the image-mask pairs $\langle \bX_t, \bhY_{t-1} \rangle \in \mathbb{R}^{H \times W \times (3+1)}$ and $\bM_n$ to a shared low-dimensional space. Then, feature matching using global convolutions~\cite{peng2017} between both features is performed and fed into the decoder part of $\phi$ to produce the estimated mask $\bhY_{t}$. In other words, we train $\phi$ to refine a rough mask from the previous frame $t-1$ in order to estimate the mask at the current frame $t$ using a reference segmentation of the object $\bM_n$.

In order to enforce temporal consistency along time, we extend the architecture from~\cite{wug2018fast} with a ``temporal skip connection". To do so, we concatenate features in the last decoder layer of $\phi$ with features extracted by the same layer in the previous frame. To reduce the memory complexity involved, we reduce the number of channels in the previous frame feature map by a factor of $1/8$ with a $3 \times 3$ convolution making the computational cost increase negligible. Adding this connection not only provides the model with information from previous frames but also acts as a simplified model memory state similar to an RNN. Moreover, when training, the gradients of future frame predictions will directly flow into previous estimates guiding the optimization to take into account that an estimated mask at frame $t$ will have a direct impact into future ones.

\paragraph{Spatial Critic.} Partial object segmentation masks and background leaks are two of the most frequent errors in segmentation. To this end, we introduce a spatial critic network, $D_\text{S}$, trained to evaluate the semantic consistency of the pixels in the estimated mask, i.e.,\ we penalize the segmented regions that do not contain one and only one fully covered object ending in its borders without extending into the background. In the experimental section, we prove this supervision to be more informative for the model in comparison to just using the cross entropy loss, as it improves its accuracy. The structure of $D_\text{S}$ resembles that of the PatchGan~\cite{pix2pix2017} mapping the product\footnote{By an abuse of notation we perform the element-wise product on the three RGB channels of $\bX_t$.} $(\bX_t \cdot \bhY_t) \in \mathbb{R}^{H \times W \times 3}$ to an output matrix $\bS \in \mathbb{R}^{H/2^6 \times W/2^6}$ where $\bS[i,j]$ is used as a partial function to compute the \textit{earth mover's distance} (EM) between the distributions of the input overlapping patch $ij$ and the real one. This critic helps to improve the difficult task of defining the mask boundaries while enforcing the model not to produce miss-classified small segmentation blobs around the objects of interest.

\paragraph{Temporal Critic.} When tracking an object instance across a video sequence we do not only need to  have semantically coherent masks, these masks must also be consistent across time. To this end, the \textit{temporal critic} $D_\text{T}$ simultaneously evaluates the current estimate w.r.t.\ to $K-1$ neighbor frames by learning the mapping $(\bx_{t-K}^t \cdot \bhy_{t-K}^t) \rightarrow S$, where as above $S \in \mathbb{R}^{H/2^6 \times W/2^6}$ is the overlapping partial scores of a PatchGan based critic. This critic helps to learn relative deformation patterns and plausible absolute motion in the segmentation mask pixel space across time. Also, it enforces the model to generate smooth transitions across mask estimates without large noisy changes.

\subsection{Learning the Model}
The loss function we define contains three terms, namely a \textit{balanced binary cross entropy} loss to penalize pixel-wise masks errors w.r.t.\ ground-truth annotations; the \textit{spatial consistency loss} to drive the distribution of the estimates to the distribution of the training masks; and the \textit{temporal consistency loss} that penalizes  temporally non-consistent masks.

\paragraph{Balanced Binary Cross Entropy Loss.} We first define the supervised pixel-wise loss for binary classification. To take into account the imbalance between the number of pixels in the object of interest and the background, we apply the balancing strategy proposed in~\cite{xie2015holistically} originally used for contour detection. Therefore, the balanced binary cross entropy loss $\mL_\text{CE}$ for $K$ frames is given by:
\begin{align}
\mL_\text{CE}=-\frac{1}{K}\sum_{t-K}^t \sum_{j \in \bY_t} & \left [ \beta \bY_{tj} \log p(\bhY_{tj}=1) \right. \\ &\hspace{-5.4 em} \left. +(1-\beta)(1-\bY_{tj})\log p(\bhY_{tj}=0) \right ]  \nonumber
\end{align}
where $\bY_t$ is the ground truth binary mask and $\beta = |\bY_t^-|/|\bY_t|$ is the percentage of pixels not belonging to the object. 

\paragraph{Spatial Consistency Loss.}
In order to optimize the \textit{spatial critic} $D_{\text{S}}$ parameters and learn the distribution of the training data, we use the modification of the standard GAN min-max strategy game~\cite{goodfellow2014generative} proposed by WGAN-GP~\cite{arjovsky2017wasserstein}. In our initial experiments, we observed that replacing the Jensen-Shannon (JS) divergence by the continuous Earth Mover Distance resulted in a more stable training. To introduce the required Lipschitz constraint, we apply the gradient penalty proposed by~\cite{gulrajani2017improved} computed as the norm of the gradients with respect to the critic input. 
Formally, if we denote the data distribution by $P_r$, the model distribution by $P_g$, and the random interpolation distribution between masked images by $P_{\wx}$, the spatial consistency loss $\mL_\text{S}$ is given by:
{\small
\begin{align}
\mL_\text{S}=\frac{1}{K}& \sum_{t-K}^t \left [ \mathbb{E}_{\bY_t \sim P_r} \left [ D_{\text{S}}(\bX_t \cdot \bY_t) \right ]  - \mathbb{E}_{\bhY_t \sim P_g} [ D_{\text{S}}(\bX_t \cdot \bhY_t) ]  \right ] \nonumber \\ &- \frac{1}{K}\sum_{t-K}^t \lambda_{\text{gp}} \mathbb{E}_{\wx \sim P_{\wx}} \left[ ( \| \nabla_{\wx} D_{\text{S}}(\wx) \|_2-1)^2\right], 
\end{align}
}%
where $\wx$ is the random interpolation between $\langle \bX_t \cdot \bY_t, \bX_t \cdot \bhY_t \rangle$ and $\lambda_{\text{gp}}$ is the penalty coefficient. 

\paragraph{Temporal Consistency Loss.}
With the previously defined losses, the segmentation regressor $\phi$ is enforced to estimate pixel-wise spatial-consistent masks. However, there is no constraint to guarantee temporal consistency,  meaning that the predicted masks should cover similar content across frames. With the \textit{temporal critic} $D_\text{T}$, we push $\phi$ to maintain temporal consistency by enforcing similarity between joint distributions of $K$ estimated and annotated masks. To estimate the distance between the distributions, we use the approximated Kantorovich-Rubinstein duality~\cite{villani2008optimal} of the Earth Mover Distance as proposed in~\cite{gulrajani2017improved}: 
\begin{align}
\label{eq:lt}
\mL_{\text{T}}= & \mathbb{E}_{x \sim P_r} \left[ D_{\text{T}}(x) \right] - \mathbb{E}_{\hx \sim P_g} [ D_{\text{T}}(\hx) ]  \nonumber \\ &- \lambda_{\text{gp}} \mathbb{E}_{\wx \sim P_{\wx}} \left[ ( \| \nabla_{\wx} D_{\text{T}}(\wx) \|_2-1)^2\right], 
\end{align}
where $x = \bx_{t-K}^t \cdot \by_{t-K}^t$ and  $\hx = \bx_{t-K}^t \cdot \bhy_{t-K}^t$ are the real and estimated conditional distributions respectively, and $\wx$ is the random interpolation between $\langle x, \hx \rangle$. Note  that, again, by an abuse of notation, we extended the element-wise product  between each $\bx_t$ and $\by_t$ (or $\bhy_t$)  along the 3 color channels of $\bx_t$.

\paragraph{Overall Loss.}
To learn to track an object instance across time, we finally define the following minmax problem:
\begin{equation}
\label{eq:fullloss}
\phi^\star = \arg \min_{\phi} \max_{D \in \mathcal{D}} \left ( \lambda_\text{CE} \mL_\text{CE} + \lambda_\text{S} \mL_\text{S} + \lambda_\text{T} \mL_\text{T} \right )
\end{equation}
where $\lambda_\text{CE}$, $\lambda_\text{S}$ and $\lambda_\text{T}$ are the hyper-parameters that control the relative importance of every loss term and $\mathcal{D}$ the set of 1-Lipschitz functions.

\section{Training Details}
Our model's encoder $\phi_\text{E}$ is a ResNet 50~\cite{he2016deep} pretrained on ImageNet~\cite{deng2009imagenet} on the task of image labeling. In order to obtain our final model, we divide the training process in two steps. 
First, our model is trained only using the supervised loss $\mL_\text{CE}$ to obtain $\phi^\star = \min_{\phi} \mL_\text{CE}$ for 6 epochs on YouTube-VOS~\cite{xu2018youtube}. 
As a result, $\phi^\star$ has an initial understanding of the video object segmentation task and provides a better initialization than ImageNet.

Then, we use $\phi^\star$ as an initialization to train our spatio-temporal model using the loss defined in Eq.~\ref{eq:fullloss} in DAVIS17~\cite{Pont-Tuset_arXiv_2017} for 40 epochs.
In our experiments, we observe that adding the critics once the model is initialized closer to the final task helps to stabilize training. 
With the idea to bring the predicted and the ground truth masks distributions in the discriminators closer at each iteration, we overwrite the ground truth pixel values $\bY_t$  with the values of the predicted masks $\bhY_t$ that are correctly estimated with an uncertainty lower than 0.25. 
Also, at each iteration, the ground truth masks are augmented by adding Gaussian noise with mean and variance equal to $\bhY_t$ statistics.

Our model is trained with images of size $512 \times 512$ augmented with horizontal flipping, random scaling with factors $[0.75, 1.25]$ and $[-30, 30]$ degrees rotations. 
Also, at each training iteration, the reference object frame $\bM_n$ of a sequence $\bx$ is randomly chosen (instead of always being the first frame in which the object appears). 
We use Adam~\cite{kingma2014adam} with a learning rate 1e-5, $\beta_1$ 0.5, $\beta_2$ 0.999, batch size 6 and polynomial decay with power 0.9. 
During the training of the spatio-temporal model, the learning rate is constant for the first 10 epochs and $\phi$ is optimized once for every 5 optimization steps of the critic networks. 
The weight coefficients for the loss terms in Eq.~\eqref{eq:fullloss} are set to $\lambda_\text{CE}=100$, $\lambda_\text{S}=1$, $\lambda_\text{T}=1$ and $\lambda_\text{gp}=10$.

In order to better approximate the mask error propagation that occurs at test time during training, we set the temporal window size $K$ to the highest value that fits in our GPU memory, $K=4$.
Note that $K$ is just used during training and only information from the previous frame is used at test time.  
This parameter is similar to \textit{back propagation through time} introduced in~\cite{hu2017maskrnn} where it was shown that training robustness improves by propagating as many $K$ estimated masks as possible rather than the ground-truth.

We concentrate all computational load to the training stage, which requires 4 NVidia\textsuperscript{\textregistered} Titan Xp , 3 of them used for training the regressor and 1 for the critics. Our model takes 2 days to finish pretraining on YouTube-VOS and 3 days for the final training on DAVIS17. During test, we only require one single GPU with at least 600Mb of memory.  When using an NVidia\textsuperscript{\textregistered} Titan Xp, we can process videos up to 32 fps.

\section{Experimental Evaluation}
\label{sec:experiments}

We thoroughly evaluate our method, \nours{}, quantitatively and qualitatively.  
We compare our approach against current state of the art on semi-supervised video object segmentation: PReMVOS~\cite{luiten2018premvos}, OSVOS\textsuperscript{S}~\cite{Man+18b}, DyeNet~\cite{Li_2018_ECCV}, CRN~\cite{Hu_2018_CVPR}, MoNet~\cite{Xiao_2018_CVPR}, RGMP~\cite{wug2018fast}, MaskRNN~\cite{hu2017maskrnn}, VideoMatch~\cite{Hu_2018_ECCV}, \nytvos{}~\cite{Xu_2018_ECCV}, PML~\cite{Chen_2018_CVPR}, \nosmn{}~\cite{Yang_2018_CVPR} and BVS~\cite{Maerki_CVPR_2016}. 

We evaluate our method on the tasks of single object (DAVIS16~\cite{Perazzi2016}) and multiple object video segmentation (DAVIS17~\cite{Pont-Tuset_arXiv_2017}, YouTube-VOS~\cite{xu2018youtube}). 
The segmentation accuracy is reported as region similarity (intersection over union $\mJ$), contour accuracy ($\mF$ measure), and their mean ($\mJ \text{\&} \mF$). For the subsets whose annotation is non-public, we compute the results using the submission website provided by the organizers of the challenge.

\subsection{Ablation Study}

In Table~\ref{table:ablation}, we perform a comprehensive ablation study to analyze the effect of the different loss components that we use in our method during training. 
In our baseline, we only use the balanced binary cross entropy loss $\mL_\text{CE}$ that penalizes wrong predictions at each pixel and frame independently. Therefore, we do not enforce any spatial or temporal consistency.

First, we introduce the spatial discriminator with its associated loss ($\mL_\text{S}$). This improves substantially the accuracy of the method in the contours of the objects boosting the $\mathcal{F}$ measure by more than 1.5 points. This improvement can also be seen qualitatively in Figure~\hyperref[1]{\ref{sec:intro}} comparing (b) to (c).

After that, we replace the previous discriminator by the temporal one with its associated loss ($\mL_\text{T}$). As a consequence, performance increases considerably gaining 1.5 points in $\mathcal{J}$\&$\mathcal{F}$. Now, the model predicts masks with better temporal consistency as can be seen in Figure~\hyperref[1]{\ref{sec:intro}} when comparing the right arm of the man in (\hyperlink{page.1}{c}) versus (\hyperlink{page.1}{d}). However, the temporal smoothness of the masks enforced by the model is sometimes too severe and the regressor has difficulties recovering from large disoccluded parts (Figure~\hyperlink{page.1}{1d}).

Finally, we combine the spatial and temporal discriminators and their respective losses. As a result, \nours{} is capable of incorporating the accuracy improvements in the contours introduced by the spatial discriminator together with the temporal masks propagation smoothness introduced by the temporal discriminator achieving a final score of $81.9$ $\mJ$\&$\mF$ in DAVIS16. As it can be seen in Figure~\hyperlink{page.1}{1e}, the final mask predicted by our spatio-temporal model segments properly the right hand of the man and it can also recover the segmentation of the legs that were occluded in previous frames.

\begin{table}[t!]
\setlength{\tabcolsep}{4pt} %
\centering
\rowcolors{5}{white}{rowblue}
\resizebox{0.25\textwidth}{!}{%
\sisetup{detect-weight=true}
\begin{tabular}{ccSSS}
\toprule
& & \multicolumn{3}{c}{DAVIS16 Val} \\
\cmidrule(lr){3-5} 
$\mL_\text{S}$ & $\mL_\text{T}$ & $\mathcal{J}$\&$\mathcal{F}$ & $\mathcal{J}$ & $\mathcal{F}$  \\
\cmidrule(lr){1-2} \cmidrule(lr){3-5}
	      -         &- 	  &          80.0 &          79.8 &          80.2 \\
		\checkmark &- 	  &          80.5 &          79.2 &          81.8 \\
		 -         &\checkmark &          81.5 &          \bfseries 80.7 &          82.2 \\
		\checkmark &\checkmark  &          \bfseries 81.9 &          80.2 &          \bfseries 83.5 \\
\bottomrule
\end{tabular}
}
\vspace{-2mm}
\caption{\textbf{Quantitative Ablation Study}: Comparison between the different loss components.}
\label{table:ablation}
\vspace{-3mm}
\end{table}

\subsection{Evaluation on DAVIS16}
\begin{table*}[t!]
\setlength{\tabcolsep}{4pt} %
\centering
\footnotesize
\rowcolors{1}{white}{rowblue}
\resizebox{\textwidth}{!}{%
\sisetup{detect-weight=true}
\begin{tabular}{clSSSSSSSSSSSS|S}
\toprule
\multicolumn{2}{c}{Measure} & \si{\nnours} & \si{\nrgmp} & \si{\nosmn} & \si{\npml} & \si{\nvideomatch} & \si{\ncrn} & \si{\ndyenet} & \si{\nytvos} & \si{\nmaskrnn} & OSVOS$^\textrm{S}$ & \si{\nmonet} & \si{\npremvos} & RGMP$^\ast$\\
\midrule
\cellcolor{rowblue}$\mathcal{J \& F}$ & Mean $\mathcal{M} \uparrow$   &   81.9  &   81.8 &   73.5 &   77.4 &   {--} &   85.0 &   {--} &   {--} &   81.3 &   86.6 &   84.8 &\bfseries 86.8 &   79.0\\
\hline
\cellcolor{white} Frames & \hspace{-1em} per second \hspace{0.05em} $\downarrow$  &\multicolumn{1}{c}{\bfseries 32.5$^\ddagger$}\hspace{-0.5 em}/32.2$^{\star}$/30.3$^{\dagger}$   &   7.7$^\star$ &   7.1$^{\diamond}$ &   3.6 &   3.1$^{\dagger}$  &   1.4$^{\dagger}$  &   0.43$^{\ddagger}$ &   0.11 &   0.11 &   0.09$^{\dagger}$ &   0.07$^{\dagger}$  &   0.06 &  32.2\hspace{-0.5 em}$^{\star}$\\
\hline
	                         & Mean $\mathcal{M} \uparrow$     &   80.2 &   81.5 &   74.0 &   75.5 &   81.0 &   84.4 &\bfseries 86.2 &   79.1 &   80.4 &   85.6 &   84.7 &   84.9 &   78.4  \\
\cellcolor{rowblue}$\mathcal{J}$ & Recall $\mathcal{O} \uparrow$   &   94.6  &   91.7 &   87.6 &   89.6 &   {--} &\bfseries 97.1 &   {--} &   {--} &   96.0 &   96.8 &   96.8 &   96.1 &   92.1 \\
	                         & Decay $\mathcal{D} \downarrow$  &   9.6  &   10.9 &   9.0 &   8.5 &   {--} &   5.6 &   {--} &   {--} &   4.4 &   5.5 &   6.4 &   8.8 &\bfseries 3.6\\
	                         & Mean $\mathcal{M} \uparrow$     &   83.5  &   82.0 &   72.9 &   79.3 &   {--} &   85.7 &   {--} &   {--} &   82.3 &   87.5 &   84.8 &\bfseries 88.6 &   79.7\\
\cellcolor{white}$\mathcal{F}$ & Recall $\mathcal{O} \uparrow$   &   94.3  &   90.8 &   84.0 &   93.4 &   {--} &   95.2 &   {--} &   {--} &   93.2 &\bfseries 95.9 &   94.7 &   94.7 &   90.8\\
	                         & Decay $\mathcal{D} \downarrow$  &   9.1  &   10.1 &   10.6 &   7.8 &   {--} &   5.2 &   {--} &   {--} &   8.8 &   8.2 &   8.6 &   9.8 &\bfseries 3.6\\
\bottomrule
\end{tabular}
}
\vspace{-2mm}
\caption{\label{tab:evaltable}\textbf{DAVIS16 Val}: \ours{} versus the most recent state of the art, more methods can be found in the DAVIS website$^2$. RGMP$^\ast$ is pretrained on YouTube-VOS instead of simulated data. Frames per second reported on a Titan X for $\dagger$, Titan Xp for $\ddagger$, Quadro M600 for $\diamond$ or 1080Ti for $\star$, methods without specifier did not report hardware in their publications.}
\end{table*}%

\begin{figure}[b!]
\centering
\resizebox{\linewidth}{!}{\begin{tikzpicture}[/pgfplots/width=1\linewidth, /pgfplots/height=0.68\linewidth, /pgfplots/legend pos=south east]
    \begin{axis}[ymin=55,ymax=90,xmin=0.04,xmax=100,enlargelimits=false,
        xlabel= Frames per second (fps),
        ylabel=Accuracy ($\mJ$\&$\mF$),
		font=\scriptsize,
        grid=both,
		grid style=dotted,
        xlabel shift={-2pt},
        ylabel shift={-5pt},
        xmode=log,
        legend columns=1,
        legend style={font=\tiny},
        legend style={/tikz/every even column/.append style={column sep=3mm}},
        minor ytick={0,0.025,...,1.1},
        ytick={0,10,...,110},
		yticklabels={0,10,20,30,40,50,60,70,80,90,100},
	    xticklabels={0,0.1,1,10,100},
        legend pos= outer north east
        ]

		\addplot+[red,mark=diamond*,only marks,line width=0.9] %
		coordinates{(32,81.9)};
        \addlegendentry{\nnours}
        \label{fig:qual_vs_time:ofl}

        \addplot+[black,mark=square*, mark size=1.3,only marks, mark options={fill=black}] 
        coordinates{(0.09,86.6)};
        \addlegendentry{\nosvoss}
        
        \addplot+[orange,mark=square*, mark size=1.3,only marks, mark options={fill=red}] 
        coordinates{(0.43,86.2)};
        \addlegendentry{\ndyenet}
        
        \addplot[green,mark=square*,only marks,line width=0.75] 
        coordinates{(3.1,81.0)};
        \addlegendentry{\nvideomatch}
        
        \addplot[blue,mark=+,only marks,line width=0.75] 
        coordinates{(1.36,85.0)};
        \addlegendentry{\ncrn}
        
        \addplot[orange,mark=+,only marks,line width=0.75] 
        coordinates{(0.07,84.8)};
        \addlegendentry{\nmonet}
        
        \addplot[green,mark=+,only marks,line width=0.75] 
        coordinates{(7.69,81.8)};
        \addlegendentry{\nrgmp}

        \addplot[blue,mark=triangle*,only marks,line width=0.75] 
        coordinates{(0.11,81.35)};
        \addlegendentry{\nmaskrnn}

        \addplot[orange,mark=triangle*,only marks,line width=0.75] 
        coordinates{(0.06,86.8)};
        \addlegendentry{\npremvos}
        
        \addplot[green,mark=diamond*,only marks,line width=0.75] 
        coordinates{(0.11,79.1)};
        \addlegendentry{\nytvos}
        
        \addplot[green,mark=halfdiamond*,only marks,line width=0.75] 
        coordinates{(3.57,77.4)};
        \addlegendentry{\npml}
        
        \addplot[blue,mark=halfdiamond*,only marks,line width=0.75] 
        coordinates{(7.14,73.5)};
        \addlegendentry{\nosmn}
        
        \addplot[orange,mark=square*,only marks,line width=0.75] 
        coordinates{(2.7,59.4)};
        \addlegendentry{\nbvs}
        
    \end{axis}
\end{tikzpicture}}
\vspace{-6mm}
   \caption{\textbf{Accuracy versus speed.} $\mJ \text{\&} \mF$ in DAVIS16 with respect to frames per second (fps).}
   \label{fig:qual_vs_time}
   \vspace{-3mm}
\end{figure}
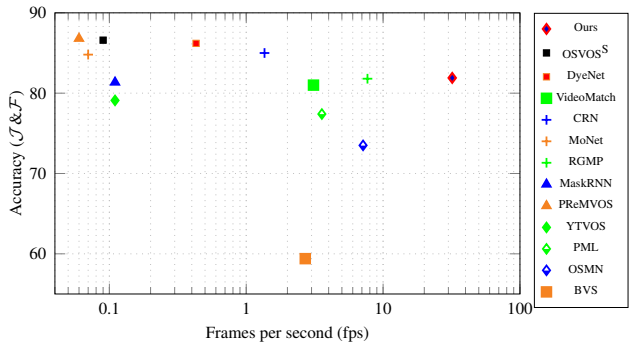

Figure~\ref{fig:qual_vs_time} shows the accuracy using $\mJ \text{\&} \mF$ in DAVIS16 versus the frame rate for various state-of-the-art methods. We can clearly see that our method is in a unique position achieving similar accuracy to previous methods that focused on speed like RGMP, OSMN or PML, while improving on their frame rate by at least a factor of $4$. When compared to methods that try to maximize their accuracy (OSVOS$^S$, MoNet or PReMVOS), we lose around $5$ points in $\mJ \text{\&} \mF$, but in exchange we increase on their frame rate by a factor of at least $350$. As a result, \nours{} sets new state of the art in terms of high frame rate while obtaining high accuracy. This bringings the field of video object segmentation closer to applications in real time scenarios.

Table~\ref{tab:evaltable} reports the results presented in Figure~\ref{fig:qual_vs_time} quantitatively and in more depth. In general, methods usually focus either on accuracy or speed at the expense of achieving lower performance in the other. Our method clearly focuses on speed while trying to retain as much accuracy as possible. Compared to the best previous method in the fast speed spectrum, RGMP, we achieve a similar accuracy while improving their frame rate  by almost a factor of 4 when tested in the same hardware (32.2 vs 7.7 fps). This improvement is mainly achieved by simplifying their multiscale testing approach by using only a single scale with an image resolution of 512x512. Note that by testing RGMP with our single scale strategy, a similar frame rate is obtained but their accuracy drops by 5 points in $\mJ \text{\&} \mF$ which is much lower than our accuracy (76.78 vs 81.9).

In order to show the effect of YouTube-VOS pretraining in previous methods, we train RGMP by substituting their synthetic data generation pretraining with YouTube-VOS video sequences, denoted as RGMP$^*$ in Table~\ref{tab:evaltable}. To do so, we first train their method on YouTube-VOS and then we fine-tuned it on DAVIS17, we stop training in both cases when the loss flattens. In order to provide a fair comparison, 
our single scale test strategy is used which also improves their fps. Pretrainig on YouTube-VOS improves their performance by roughly 2 points in $\mJ \text{\&} \mF$ (79.0 vs 76.78) which is sill significantly lower than the accuracy of our model.

\addtocounter{footnote}{1}
\footnotetext{\scriptsize{\url{https://davischallenge.org/davis2016/soa_compare.html}}}

\begin{table}[t!]
\setlength{\tabcolsep}{4pt} %
\centering
\footnotesize
\rowcolors{5}{white}{rowblue}
\resizebox{\linewidth}{!}{%
\sisetup{detect-weight=true}
\begin{tabular}{cS[table-format=2.1]cS[table-format=2.1]S[table-format=2.1]S[table-format=2.1]cS[table-format=2.1]S[table-format=2.1]}
\toprule
 & \multicolumn{4}{c}{DAVIS17 Val} & \multicolumn{4}{c}{YouTube-VOS Val} \\
 \cmidrule(lr){2-5} \cmidrule(lr){6-9}
 & $\mathcal{J}$\&$\mathcal{F}$ & $fps$ & $\mathcal{J}$ & $\mathcal{F}$ & $\mathcal{J}$\&$\mathcal{F}$ & $fps$ & $\mathcal{J}$ & $\mathcal{F}$ \\
\cmidrule(lr){1-9}
        \nnours   & 60.2 &\textbf{16.5$^\ddagger$}/16.3$^\star$/15.4$^\dagger$ & 57.6 & 63.4 & 52.0 &\textbf{17.2$^\ddagger$}/17.1$^\star$/16.1$^\dagger$ & 50.0 & 53.9 \\
        \nrgmp	  &	66.7 & 2.97$^\star$ & 64.8 & 68.6 & 52.8 & 2.66$^\star$ & 51.4 & 54.2 \\
        \nosmn	  &	54.8 & 3.62$^\diamond$ & 52.5 & 57.1 & 51.8 & 3.79$^\diamond$ & 50.3 & 52.1 \\
		OSVOS$^\textrm{S}$  & 68.0 & 0.05$^\dagger$ & 64.7 & 71.3 & {--} & {--} & {--} & {--} \\
		\npremvos &\bfseries 77.8 & 0.03 &\bfseries 73.9 &\bfseries 81.8 &\bfseries 72.2 & 0.03 &\bfseries 69.3 &\bfseries 75.2 \\
		\cmidrule(lr){1-9}
		RGMP$^\ast$  &	56.2 & 16.3$^\star$ & 52.8 & 59.6 & 46.3 & 17.1$^\star$ & 44.3 & 48.2 \\

\bottomrule
\end{tabular}
}
\vspace{-2mm}
\caption{\label{tab:evaltable2}\textbf{DAVIS17 and YouTube-VOS}: \ours{} versus the state of the art, more methods can be found in the DAVIS website$^3$. $fps$ stands for frames per second and is computed assuming linear scaling with the number of objects, thus using $fps$ from DAVIS16 and multiplying by the mean number of objects in a certain set. Specifier ($\ast$, $\dagger$, $\ddagger$, $\diamond$, $\star$) definitions are the same than in Table~\ref{tab:evaltable}.}
\vspace{-3mm}
\end{table}
\stepcounter{footnote}\footnotetext{\scriptsize{\url{https://davischallenge.org/davis2017/soa_compare.html}}}

\subsection{Evaluation on DAVIS17 and YouTube-VOS}
We also report the performance of \nours{} in multi-object video segmentation in Table~\ref{tab:evaltable2} against state of the art methods in DAVIS17 and YouTube-VOS. Our method is still the fastest at such accuracy outperforming other competitors that are 4 times slower. When compared to RGMP pretrained in YouTube-VOS and tested with our single scale strategy, for the same speed we outperform their model by 4 and 5.7 points in $\mathcal{J}$\&$\mathcal{F}$ in both DAVIS17 and YouTube-VOS, respectively.

\begin{figure*}[t]
\centering
\resizebox{0.98\textwidth}{!}{%
	  \setlength{\fboxsep}{0pt}
      \rotatebox{90}{\hspace{6.2mm}193aa74f36}
      \fbox{\includegraphics[width=0.3\textwidth]{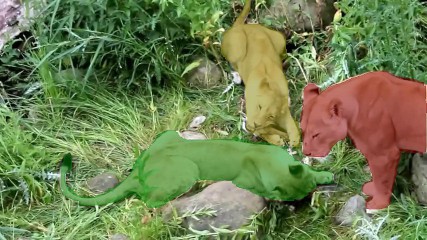}}   
      \fbox{\includegraphics[width=0.3\textwidth]{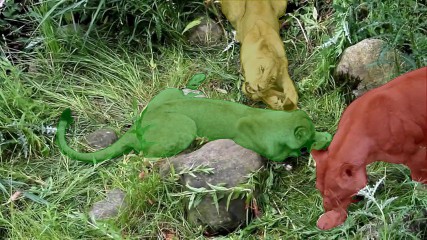}}
      \fbox{\includegraphics[width=0.3\textwidth]{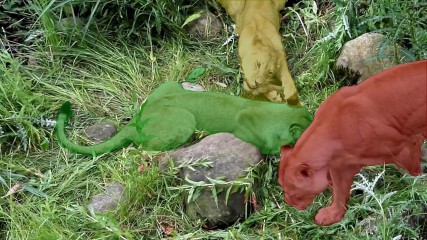}}
      \fbox{\includegraphics[width=0.3\textwidth]{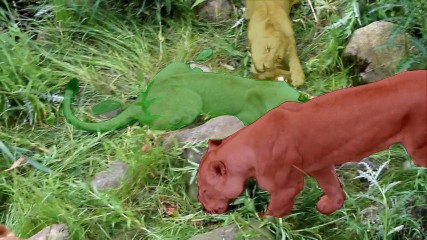}}
      \fbox{\includegraphics[width=0.3\textwidth]{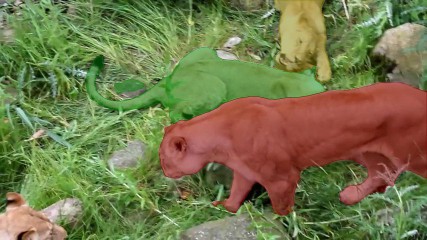}}
      }\\[1mm]
\resizebox{0.98\textwidth}{!}{%
	  \setlength{\fboxsep}{0pt}
      \rotatebox{90}{\hspace{5.7mm}19904980af}
      \fbox{\includegraphics[width=0.3\textwidth]{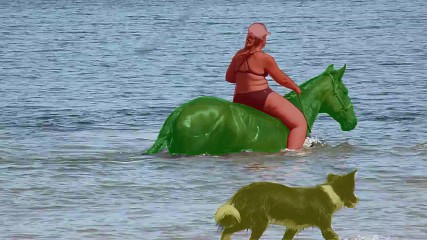}}
      \fbox{\includegraphics[width=0.3\textwidth]{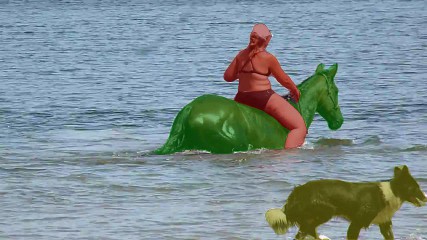}}
      \fbox{\includegraphics[width=0.3\textwidth]{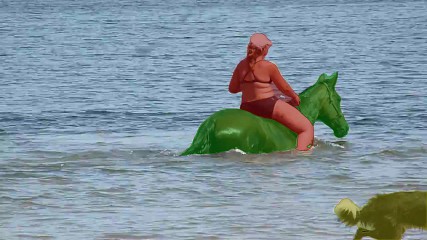}}
      \fbox{\includegraphics[width=0.3\textwidth]{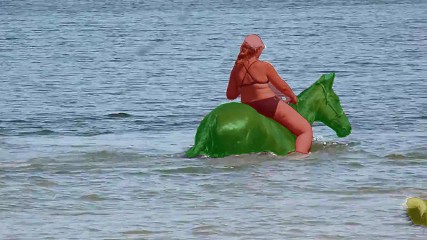}}
      \fbox{\includegraphics[width=0.3\textwidth]{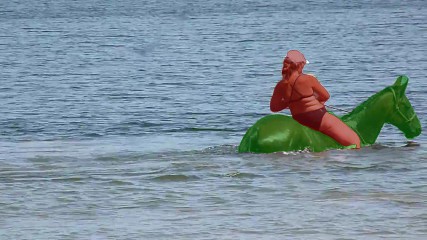}}
      }\\[1mm]
\resizebox{0.98\textwidth}{!}{%
	  \setlength{\fboxsep}{0pt}
      \rotatebox{90}{\hspace{6.5mm}Breakdance}
      \fbox{\includegraphics[width=0.3\textwidth]{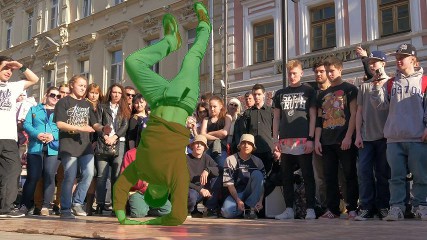}}   
      \fbox{\includegraphics[width=0.3\textwidth]{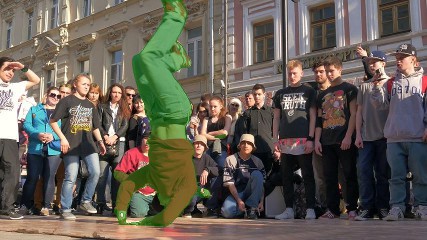}}
      \fbox{\includegraphics[width=0.3\textwidth]{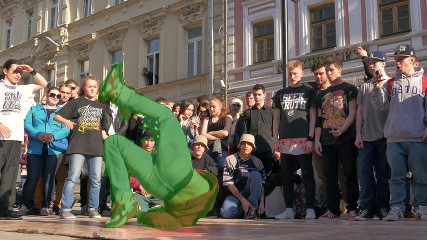}}
      \fbox{\includegraphics[width=0.3\textwidth]{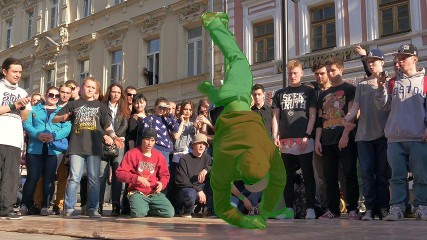}}
      \fbox{\includegraphics[width=0.3\textwidth]{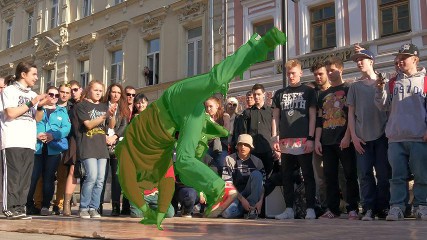}}
      }\\[1mm]
\resizebox{0.98\textwidth}{!}{%
	  \setlength{\fboxsep}{0pt}
      \rotatebox{90}{\hspace{11mm}Judo}
      \fbox{\includegraphics[width=0.3\textwidth]{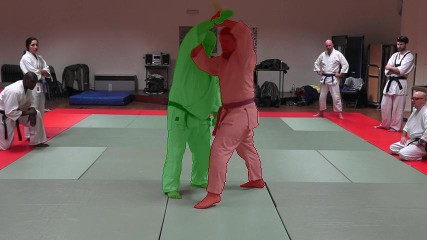}}   
      \fbox{\includegraphics[width=0.3\textwidth]{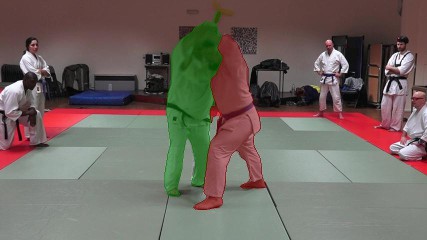}}
      \fbox{\includegraphics[width=0.3\textwidth]{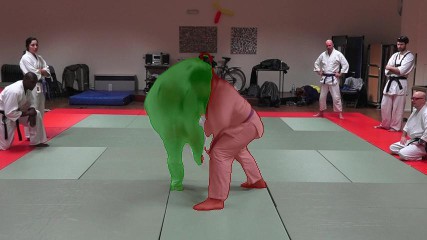}}
      \fbox{\includegraphics[width=0.3\textwidth]{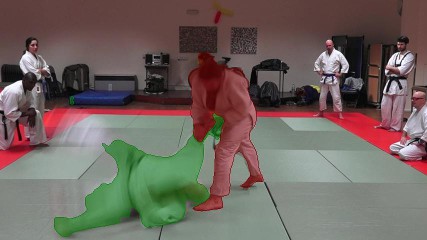}}
      \fbox{\includegraphics[width=0.3\textwidth]{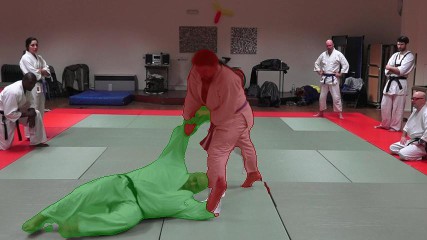}}
      }\\[1mm]
\resizebox{0.98\textwidth}{!}{%
	  \setlength{\fboxsep}{0pt}
      \rotatebox{90}{\hspace{10mm}Libby}
      \fbox{\includegraphics[width=0.3\textwidth]{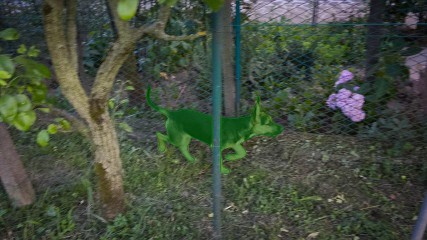}}   
      \fbox{\includegraphics[width=0.3\textwidth]{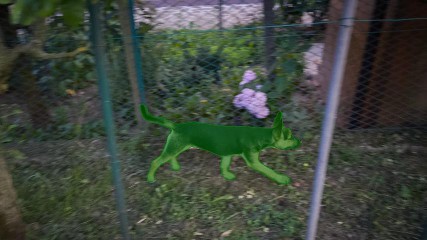}}
      \fbox{\includegraphics[width=0.3\textwidth]{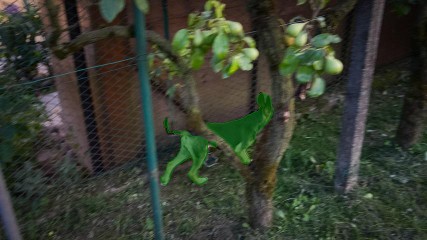}}
      \fbox{\includegraphics[width=0.3\textwidth]{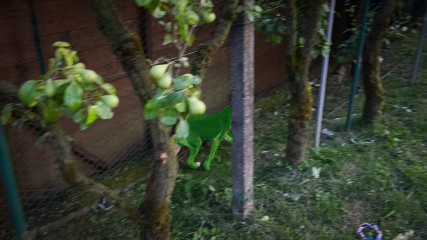}}
      \fbox{\includegraphics[width=0.3\textwidth]{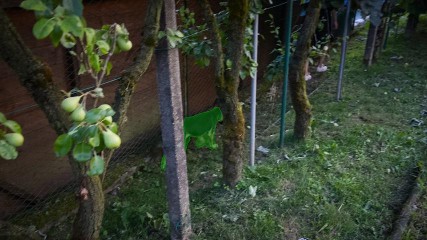}}
      }\\[1mm]
\resizebox{0.98\textwidth}{!}{%
	  \setlength{\fboxsep}{0pt}
      \rotatebox{90}{\hspace{5mm}Drift-Chicane}
      \fbox{\includegraphics[width=0.3\textwidth]{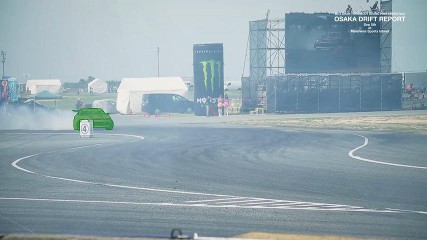}}   
      \fbox{\includegraphics[width=0.3\textwidth]{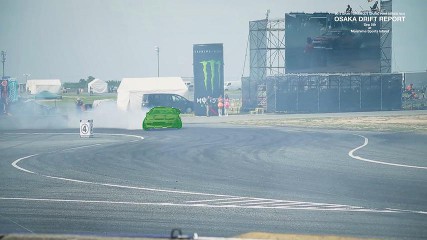}}
      \fbox{\includegraphics[width=0.3\textwidth]{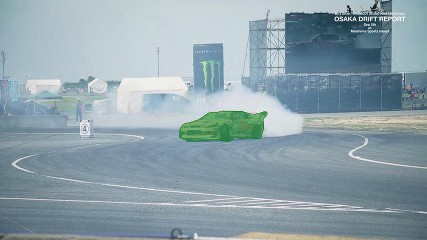}}
      \fbox{\includegraphics[width=0.3\textwidth]{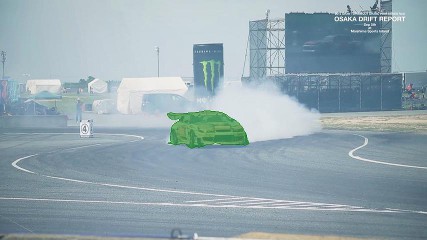}}
      \fbox{\includegraphics[width=0.3\textwidth]{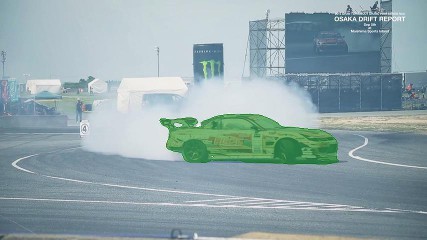}}
      }\\[1mm]
\caption{\textbf{Qualitative results}: Sample sequences from YouTube-VOS (top 2), DAVIS17 (middle 2) and  DAVIS16 (bottom 2). Leftmost image is the initial reference frame, the rest of the images are the predictions for the following frames.}
\label{fig:qualitative}
\vspace{-2mm}
\end{figure*}

Against the original RGMP model, our method has lower accuracy in DAVIS17, but it has similar results in YouTube-VOS. In DAVIS17, there are several small objects and their multi-scale testing strategy helps to obtain better performance in such a scenario. Moreover, their model in the multi-object scenario not only predicts the masks for objects in the sequence, but also propagates the background as an additional object increasing the average number of objects. As a result of not using multi-scale testing and not propagating an additional object, \nours{} runs more than 5 times faster than their method.

Compared to the best performing method, PReMVOS, our inference speed is more than 500 times faster. However, achieving such high frame rate comes at the cost of a drop in accuracy. We believe that optimizing the inference while achieving high accuracy is more challenging in the multi-object scenario compared to the single object one and poses an interesting direction for future works.

\vspace{2mm}
\subsection{Fairness in method comparison}
We would like to briefly discuss the difficulties in providing a fair comparison with other video object segmentation models. First of all, methods in Table~\ref{tab:evaltable} and Table~\ref{tab:evaltable2} have been pretrained in a wide variety of different datasets. For instance, OSVOS$^S$, DyNet, OSMN, PML and PReMVOS use COCO~\cite{Lin2014}; MoNet, PReMVOS, DyeNet, VideoMatch, PML, RGMP and CRN use PASCAL VOC~\cite{Everingham2010, Hariharan2011}; and RGMP uses as well ECSSD~\cite{Shi2016} and MSRA 10K~\cite{Cheng2015}.
Also, MoNet, PReMVOS, DyeNet and CRN use optical flow produced by Flownet2.0~\cite{Ilg2017} which is trained using~\cite{Dosovitskiy2015,Mayer2016}. Before the release of YouTube-VOS, static image datasets were used to train most methods due to the lack of a large scale video object segmentation dataset. We expect future methods in the field to gradually converge to YouTube-VOS pretraining which would make the comparison among different methods easier.

Moreover, previous methods report timings in a wide variety of GPU types. In order to provide a fair comparison, we list the GPU type used in each publication in Table~\ref{tab:evaltable} and and Table~\ref{tab:evaltable2} when available and we test our method in the three different GPU types that we have at our disposal, 1080Ti, Titan Xp and, Titan X.

\vspace{2mm}
\subsection{Qualitative Results}
Figure~\ref{fig:qualitative} shows examples of the predicted masks using our approach, \nours{}. The first column displays the reference mask $\bM_n$ and the rest of the columns display the segmented mask by our method in the following frames. 
\vspace{3mm}

Note that even when the input frame is corrupted by occlusions, changes of appearance and dynamic background, our method remains robust.

\section{Conclusions}
We have presented \nours{}, which, to the best of our knowledge, is the first real-time approach for semi-supervised video object segmentation running at 32 fps and yielding high-quality segmentation masks. \nours{}'s accuracy is on a par with previous state of the art optimized for speed but it runs at a much higher frame rate. 

To achieve this, we have designed a novel GAN architecture made of a relatively small regressor and two critics that enforce spatio-temporal consistency over finite temporal windows during training. 
At test time, the critics are removed, leading to a simple but robust regressor that does not require fine-tuning nor post-processing operations when applied to new sequences with unseen objects. This  opens a wide range of potential applications in the near future for real-time video analysis and video editing.

\section*{Acknowledgments}
The authors would like to thank Fabian Mentzer and Kevis-Kokitsi Maninis for the insightful discussions and proofreading of this manuscript. 
This work is supported in part by the Swiss Commission for Technology and Innovation (CTI, Grant No. 19015.1 PFES-ES, NeGeVA), by the Spanish Ministry of Science and Innovation under projects HuMoUR TIN2017-90086-R, ColRobTransp DPI2016-78957 and Mar\'ia de Maeztu Seal of Excellence MDM-2016-0656; and by the EU project AEROARMS ICT-2014-1-644271. We also thank NVidia for hardware donation under the GPU Grant Program.

{\small
\bibliographystyle{ieee}
\bibliography{egbib}
}

\end{document}